\documentclass[fleqn,10pt]{wlscirep}
\usepackage[utf8]{inputenc}
\usepackage[T1]{fontenc}
\usepackage{subcaption}
\usepackage{multirow}
\usepackage{threeparttable}
\usepackage{array}
\usepackage{supertabular}
\usepackage{algorithm}
\usepackage{algorithmicx}
\usepackage{algpseudocode}
\usepackage{caption}
\usepackage{hyperref}
\usepackage[cal=cm]{mathalfa}

\usepackage[switch]{lineno} 

\usepackage{soul}
\usepackage{color}
\soulregister\cite7
\soulregister\citep7
\soulregister\citet7
\soulregister\ref7
\soulregister\pageref7

\begin{document}


\title{Cross-Linguistic Persona-Driven Data Synthesis for Robust Multimodal Cognitive Decline Detection}

\author[1,$\dagger$]{Rui Feng} 
\author[2,$\dagger$]{Zhiyao Luo} 
\author[1]{Liuyu Wu}
\author[1]{Wei Wang}
\author[3]{Yuting Song}
\author[3]{Yong Liu}
\author[4, 5]{Kok Pin Ng}
\author[1,*]{Jianqing Li}
\author[3,*]{Xingyao Wang}

\affil[1]{Engineering Research Center of Intelligent Theranostics Technology and Instruments, Ministry of Education, School of Biomedical Engineering and Informatics, Nanjing Medical University, Nanjing, 211166, China}
          
\affil[2]{Institute of Biomedical Engineering, University of Oxford, Oxford, OX1 2JD, United Kingdom}

\affil[3]{Institute of High Performance Computing, Agency for Science, Technology and Research (A*STAR), Singapore, 138632, Singapore}

\affil[4]{Department of Neurology, National Neuroscience Institute, Singapore, 308433, Singapore}
\affil[5]{Duke-NUS Medical School, Singapore, 169857, Singapore}

\affil[$\dagger$]{These authors contributed equally to this work}
\affil[*]{Corresponding author: jqli@njmu.edu.cn, wang\_xingyao@a-star.edu.sg}

\begin{abstract}
Speech-based digital biomarkers represent a scalable, non-invasive frontier for the early identification of Mild Cognitive Impairment (MCI). However, the development of robust diagnostic models remains impeded by acute clinical data scarcity and a lack of interpretable reasoning. Current solutions frequently struggle with cross-lingual generalization and fail to provide the transparent rationales essential for clinical trust. To address these barriers, we introduce SynCog, a novel framework integrating controllable zero-shot multimodal data synthesis with Chain-of-Thought (CoT) deduction fine-tuning. Specifically, SynCog simulates diverse virtual subjects with varying cognitive profiles to effectively alleviate clinical data scarcity. This generative paradigm enables the rapid, zero-shot expansion of clinical corpora across diverse languages, effectively bypassing data bottlenecks in low-resource settings and bolstering the diagnostic performance of Multimodal Large Language Models (MLLMs). 
Leveraging this synthesized dataset, we fine-tune a foundational multimodal backbone using a CoT deduction strategy, empowering the model to explicitly articulate diagnostic thought processes rather than relying on black-box predictions. 
Extensive experiments on the ADReSS and ADReSSo benchmarks demonstrate that augmenting limited clinical data with synthetic phenotypes yields competitive diagnostic performance, achieving Macro-F1 scores of 80.67\% and 78.46\%, respectively, outperforming current baseline models.
Furthermore, evaluation on an independent real-world Mandarin cohort (CIR-E) demonstrates robust cross-linguistic generalization, attaining a Macro-F1 of 48.71\%.
These findings constitute a critical step toward providing clinically trustworthy and linguistically inclusive cognitive assessment tools for global healthcare.
\end{abstract}


\flushbottom
\maketitle

\thispagestyle{empty}
\section*{Introduction}

Alzheimer’s disease (AD) is a progressive neurodegenerative disorder characterized by the gradual erosion of cognitive functions, memory, and activities of daily living. Driven by the global demographic shift toward an aging population, the prevalence of AD has escalated significantly, emerging as a critical public health priority~\cite{gustavsson2023global, better2023alzheimer}.
As a critical transitional phase, Mild Cognitive Impairment (MCI) represents a vital clinical window for the early identification of AD. Identifying individuals at this stage and implementing timely interventions are essential to slow disease progression and secure better cognitive outcomes.

Compared to standard clinical diagnoses (e.g., comprehensive neuropsychological assessments, fluid biomarker analysis, and neuroimaging techniques)~\cite{valletta2025blood, arevalo2015mini}, speech and language production has been recognized as sensitive indicators of cognitive decline~\cite{sanchez2024identification}, as the complex process of formulating speech requires the integrity of multiple cognitive domains, including episodic memory, executive function, and lexical retrieval. 
In particular, picture description tasks (e.g., the ``Cookie Theft'' task) provide a unique and clinically robust paradigm for eliciting speech~\cite{shafiyan2025revisiting}. Picture description constrains the semantic context, requiring the subject to visually process a scene and retrieve specific lexical items to describe simultaneous actions. Leveraging this paradigm, AI-based automated assessment techniques have gained significant traction. By utilizing Natural Language Processing (NLP) and acoustic analysis to evaluate recordings from picture description tasks, these systems offer a scalable, cost-effective, and objective alternative for mass screening and longitudinal monitoring~\cite{lima2025evaluating, casu2025integrating}.

Recently, Multimodal Large Language Models (MLLMs) have demonstrated transformative potential in biomedical tasks, owing to their capacity to process multimodal inputs and encode vast amounts of semantic knowledge~\cite{mcduff2025towards, du2024enhancing}, which offer a promising avenue for scalable cognitive assessment. However, deploying MLLMs for clinical MCI detection currently faces three critical challenges. 
First, data scarcity creates a significant bottleneck. High-quality speech-text MCI datasets are often limited in size due to privacy concerns and collection costs. For instance, the widely used ADReSSo dataset comprises only 237 participants~\cite{luz2021detecting}. Applying Supervised Fine-Tuning (SFT) on such small-scale datasets often leads to overfitting, where models may hallucinate non-existent linguistic cues or capture spurious correlations from background noise rather than genuine cognitive markers~\cite{huang2025survey}. 
Second, there is a lack of explicit diagnostic reasoning. Clinical MCI assessment relies on subjective and complex decision-making processes, resulting in a scarcity of standardized clinical diagnostic thought processes recorded as step-by-step reasoning annotations. This absence hinders the interpretability of AI models, rendering them black boxes that fail to engender clinical trust.
Third, cross-lingual generalization remains a critical bottleneck. Existing research is predominantly confined to English-centric datasets, resulting in a severe scarcity of multilingual clinical corpora. Consequently, models often fail to generalize across different languages and dialects, leading to significant performance degradation when deployed in diverse global populations~\cite{wang2024llm}.

\begin{figure}[t]
\centering
\includegraphics[width=0.95\linewidth]{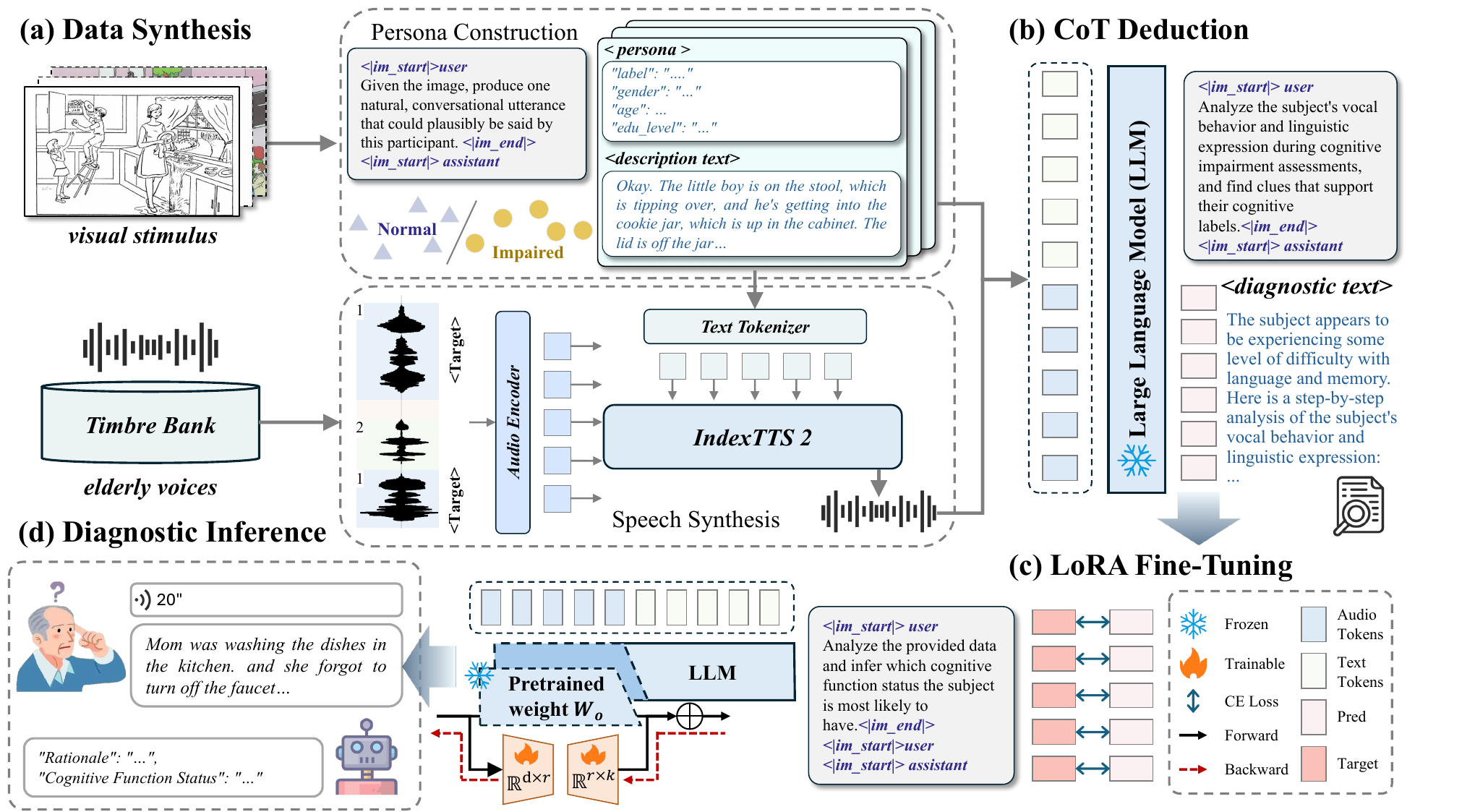}
\caption{\textbf{Overview of the proposed SynCog framework.} 
The pipeline consists of four phases: 
\textbf{(a)} Data Synthesis, which involves simulating subjects via LLMs to generate high-fidelity multimodal narratives comprising both audio recordings and their corresponding transcripts, conditioned on demographic and cognitive attributes; 
\textbf{(b)} Chain-of-Thought Distillation, where diagnostic rationales are systematically derived to bridge raw multimodal evidence with clinical labels; 
\textbf{(c)} CoT deduction Fine-Tuning, utilizing Low-Rank Adaptation (LoRA) to optimize the model $\mathcal{M}$ based on the diagnostic reasoning sequences derived during CoT Distillation; 
and \textbf{(d)} Diagnostic Inference, where the fine-tuned model $\mathcal{M}$ first articulates a heuristic reasoning sequence $r_j$ to ground the final assessment in specific pathological markers. 
By synthesizing evidence from acoustic prosody and linguistic content, this transparent inference pathway effectively mitigates the risk of shortcut learning. The resulting diagnostic process not only enhances the robustness of results across different linguistic contexts but also provides clinicians with interpretable evidentiary support to validate the final assessment.
}
\label{fig:framework}
\end{figure}

To address these challenges, we propose SynCog, a novel framework that overcomes the inherent constraints of data-scarce clinical supervision by integrating a controllable phenotypic simulation framework with a logic-driven Chain-of-Thought (CoT) deduction strategy.
Specifically, to mitigate data scarcity and imbalance, we introduce a controllable data synthesis pipeline. Unlike previous augmentation methods~\cite{wu2024improving} that operate on a single modality, our approach constructs comprehensive digital personas defined by demographic attributes and linguistic styles. We leverage MLLMs to generate diverse, persona-consistent picture description narratives and employ advanced voice cloning techniques to synthesize corresponding speech signals. This process yields a large-scale, privacy-compliant, and fully labeled multimodal dataset that mirrors the variability of real-world patient populations. 
Furthermore, to compensate for the lack of expert reasoning annotations, we integrate a CoT deduction Fine-Tuning strategy. Rather than training the model solely to predict diagnostic labels, we guide it to generate explicit CoT reasoning traces. By distilling high-quality self-generated rationales, we enforce a learning process that prioritizes diagnostic logic over shallow surface patterns.
Finally, to validate the clinical robustness and scalability of our framework, we conducted extensive experiments on both standardized English benchmarks and an independently collected real-world Mandarin clinical cohort. The results demonstrate that SynCog achieves superior performance compared to existing baselines. Crucially, our findings confirm the framework's ability to bridge the linguistic gap in cognitive assessment, proving that synthesis-augmented approaches can generalize effectively across diverse languages and clinical environments.





\section*{Results}
\label{sec:results}

\subsection*{Overview of SynCog}

To address the dual challenges of data scarcity and the lack of interpretability in AI-based cognitive assessment, we developed a novel framework that integrates generative data synthesis with CoT deduction fine-tuning. Specifically, we instantiate our framework using Qwen2-Audio-7B-Instruct~\cite{Chu2024Qwen2Audio} as the multimodal backbone. We employ this architecture due to its native audio-text alignment and robust instruction-following capabilities, which have established it as a reliable foundation for comparative multimodal research. The overall workflow is illustrated in Fig.~\ref{fig:framework}. Unlike conventional deep learning approaches that rely solely on discriminative features, our system first constructs a stratified synthetic cohort. This cohort comprises both cognitively normal and impaired digital subjects, enabling the model to learn robust pathological representations. Subsequently, through reasoning self-distillation, the model is trained to elucidate its diagnostic logic. This process maps multimodal inputs, including speech audio and transcripts, to a predicted clinical status while providing a corresponding clinical rationale.

\begin{table}[h]
\centering
\fontsize{9}{11}\selectfont\setlength{\tabcolsep}{3pt}
\begin{threeparttable}
\caption{Detailed clinical and demographic profiles of the public datasets, real-world, and synthetic cohorts.}
\label{tab:detailed_cohorts}
\begin{tabular}{lcccccccc}
\toprule
\textbf{Cohorts} & \textbf{Language} & \textbf{Group} & \textbf{Subjects ($n$)} & \textbf{Age} & \textbf{Sex ($n$)} & \textbf{Education level} & \textbf{MMSE} & \textbf{MoCA} \\
\midrule
\textit{Public Datasets} & & & & & & & & \\

\multirow{2}{*}{\textbf{ADReSS}} & \multirow{2}{*}{English} 
   & AD & 78 & 66.6 (6.6) & 35 / 43 & -- & 17.8 (5.6) & -- \\
 & & non-AD & 78 & 66.8 (6.3) & 35 / 43 & -- & 29.0 (1.2) & -- \\

\multirow{2}{*}{\textbf{ADReSSo}} &  \multirow{2}{*}{English}  
   & AD & 122 & -- & -- & -- & 17.8 (5.5) & -- \\
 & & non-AD & 115 & -- & -- & -- & 29.0 (1.2) & -- \\
\midrule
\textit{Real-world Cohort} & & & & & & & & \\
\multirow{3}{*}{\textbf{CIR-E}} & \multirow{3}{*}{Mandarin} 
  & AD & 46 & 75.1 (5.4)  & 29 / 17 & 3 / 24 / 16 / 3 & 23.2 (2.6)  & 16.3 (2.0) \\
& & MCI & 74 & 73.6 (5.4) & 44 / 30 & 3 / 18 / 51 / 2 & 27.0 (2.1) & 22.2 (2.0) \\
& & HC & 33 & 71.1 (4.2) & 18 / 15 & 2 / 5 / 23 / 3 & 28.2 (1.1) & 27.0 (1.5) \\

\midrule
\textit{Synthetic Cohorts} & & & & & & & & \\

\multirow{2}{*}{\textbf{SYN-EN}} &  \multirow{2}{*}{English} 
   & AD & 500  & 73.1 (7.2) & 250 / 250 & 192 / 164 / 104 / 40 & -- & --  \\
 & & non-AD & 500 & 73.1 (7.0) & 250 / 250 & 205 / 146 / 103 / 46 & -- & -- \\

\multirow{3}{*}{\textbf{SYN-ZH}} & \multirow{3}{*}{Mandarin} 
   & AD & 500 & 72.4 (6.8) & 250 / 250 & 189 / 162 / 102 / 47 & -- & --  \\
 & & MCI & 500 & 71.9 (7.3) & 250 / 250 & 198 / 159 / 98 / 45 & -- & -- \\
 & & HC & 500 & 72.4 (7.4) & 250 / 250 & 204 / 151 / 90 / 55 & -- & -- \\

\bottomrule
\end{tabular}

\begin{tablenotes}
        \footnotesize
        \item[1] Age, MMSE, and MoCA are reported as mean (sd).
        \item[2] Gender is reported as Female / Male counts.
        \item[3] Education: Primary or below / Junior high / High school / University or above.
    \end{tablenotes}
\end{threeparttable}
\end{table}

\subsection*{Data description}

In this study, we aim to construct multi-dimensional patient personas and generate synthetic speech data across varying cognitive levels. To this end, we utilize three datasets to comprehensively evaluate the proposed framework: two publicly available English benchmarks, ADReSS and ADReSSo, and an independently collected Mandarin corpus, CIR-E. 
The detailed clinical and demographic profiles of these cohorts, alongside their synthetic counterparts, are summarized in Table~\ref{tab:detailed_cohorts}. Leveraging these diverse data sources, which vary in diagnostic labeling schemes and disease severity, allows us to validate the cross-lingual transferability and robustness of our framework, ensuring the model accurately captures and generalizes linguistic biomarkers across diverse real-world scenarios.

\textbf{ADReSS} was introduced as part of the INTERSPEECH 2020 ADReSS Challenge~\cite{luz2020alzheimer}. It consists of speech recordings and corresponding transcripts collected from spontaneous picture description tasks based on the ``Cookie Theft'' picture from the Boston Diagnostic Aphasia Examination. The dataset includes an equal number of participants diagnosed with AD and cognitively normal controls, balanced for gender and age to minimize demographic bias. 
Each audio sample has a duration of approximately one to two minutes, with an accompanying transcript.
The ADReSS dataset is designed for binary classification (AD and non-AD) and has been preprocessed to remove personally identifiable information and background noise.

\textbf{ADReSSo} was released for the INTERSPEECH 2021 ADReSSo Challenge~\cite{luz2021detecting} as an extension of the original ADReSS dataset.
Unlike ADReSS, which includes both audio and transcripts, ADReSSo focuses solely on the acoustic modality, providing only raw audio recordings for training and evaluation. The dataset contains recordings collected in naturalistic settings, emphasizing robustness against variations in recording conditions and spontaneous speech behaviors. All recordings were normalized and manually screened to ensure consistent audio quality and speaker intelligibility.

To assess real-world generalization, we collected an independent Mandarin corpus, \textbf{CIR-E}~\cite{feng2025cogbenchlargelanguagemodel}, from community-dwelling older adults in Jiangsu, China. CIR-E captures spontaneous speech during a one-minute picture description task. This corpus comprises 153 samples with labels assigned by neurologists according to a ternary scheme of Healthy Control (HC), MCI, and AD.
For individuals meeting the preliminary screening criteria, the research protocol was explained in detail, and informed consent was obtained. Subsequently, experienced neurologists conducted a comprehensive clinical evaluation, which included standardized cognitive assessments (e.g., MMSE, MoCA) and evaluation of activities of daily living (ADL). Group inclusion was then determined based on the results of these assessments, alongside medical history and physical examination. Specifically, participants with cognitive impairment were identified through this clinical assessment, while HC were defined as individuals with no subjective memory complaints and no history of major neurological, psychiatric, or metabolic disorders. Strict exclusion criteria (e.g., other neurological or psychiatric disorders, systemic organ failure, cognition-affecting medications) were applied to ensure that observed speech characteristics primarily reflect AD-related cognitive decline.

\begin{figure}[h]
  \centering
  \includegraphics[width=0.95\linewidth]{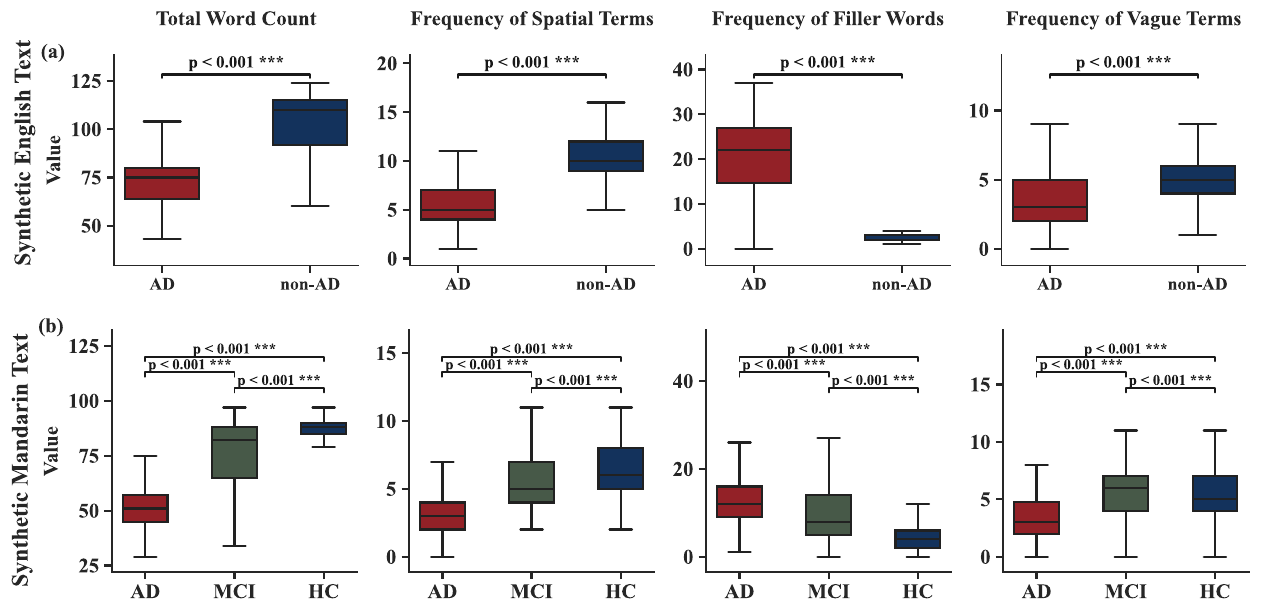}
  \caption{\textbf{Distribution of key linguistic biomarkers across synthetic cohorts generated by SynCog.} Box plots quantify four representative features mapped to the assessment dimensions and scoring criteria: Total Word Count, Frequency of Spatial Terms, Frequency of Filler Words, and Frequency of Vague Terms. These metrics characterize the linguistic profiles of the generated cohorts across the diagnostic spectrum. \textbf{(a)} Synthetic English dataset contrasting patients with Alzheimer’s disease (AD) and non-AD individuals. \textbf{(b)} Synthetic Mandarin dataset across three diagnostic categories: healthy controls (HC), mild cognitive impairment (MCI), and AD. The distinct separation between neurodegenerative groups and healthy controls confirms that the generated text preserves pathological patterns consistent with the established scoring criteria and clinical diagnostic standards.}
  \label{fig:text_features}
\end{figure}

\subsection*{Synthetic Cohorts Reproduce Clinical Linguistic and Acoustic Phenotypes}

Before assessing diagnostic performance, we rigorously validated the quality and clinical fidelity of the multimodal data generated by SynCog. To systematically investigate the scalability of our framework, we constructed stratified synthetic cohorts designated as SYN-EN for the English domain and SYN-ZH for the Mandarin domain by progressively increasing the data volume. 
Specifically, we applied five synthetic data scales, where each diagnostic category within the cohorts was expanded from 50 to 500 subjects for both the English and Mandarin cohorts.
This design yielded a maximum synthetic corpus of 1,000 subjects for the binary SYN-EN cohort (AD and non-AD) and 1,500 subjects for the ternary SYN-ZH cohort (AD, MCI, and HC).
Such proportional generation preserves a balanced distribution across all diagnostic phenotypes while enabling a systematic evaluation of model performance under increasing data scales.

\subsubsection*{Synthesis of Linguistic Biomarkers}

We analyzed key linguistic biomarkers associated with neurodegeneration to validate the clinical fidelity of our synthetic cohorts. As illustrated in Fig.~\ref{fig:text_features}, the generated data successfully recapitulates clinically established pathological trends across both English and Mandarin languages. Specifically, the synthetic AD and MCI groups exhibited a statistically significant reduction in total word count and the usage of spatial terms compared to HC ($p < 0.001$). Conversely, indicators of disfluency and anomia, quantified by the frequency of filler words and vague terms, showed a significant increase proportional to cognitive severity ($p < 0.001$). These distinct distributional shifts confirm that our phenotype modeling captures the progressive nature of linguistic impairment rather than merely generating generic descriptions.

\subsubsection*{Consistency of Acoustic Distributions}

\begin{figure}[t]
  \centering
  \includegraphics[width=0.45\linewidth]{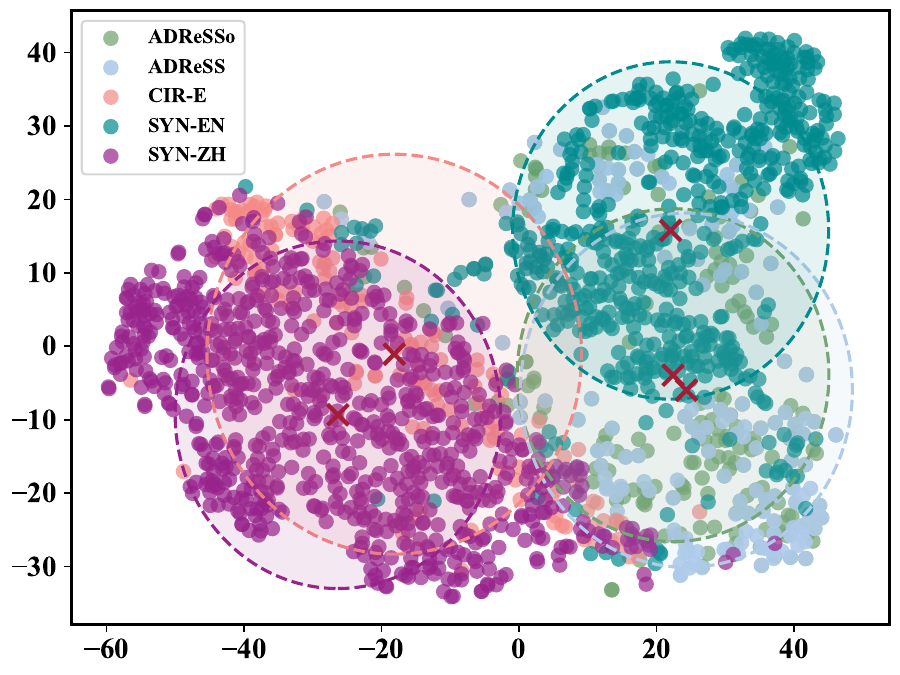}
  \caption{\textbf{Distributional alignment of synthetic and clinical acoustic embeddings.} The t-SNE visualization displays high-dimensional feature vectors extracted from speech samples using the \textit{wav2vec2-base-960h} model. The plot reveals two distinct linguistic clusters where the synthetic data shares a substantial manifold overlap with real-world recordings. The English cluster comprises the ADReSS and ADReSSo clinical baselines aligned with the synthetic English dataset. The Mandarin cluster includes the CIR-E real-world cohort aligned with the synthetic Mandarin cohort. The dashed boundaries indicate the distributional extent of each subgroup and highlight the preservation of language-specific acoustic phenotypes by the SynCog framework.}
  \label{fig:audio_tsne}
\end{figure}

In the acoustic domain, the t-SNE visualization shown in Fig.~\ref{fig:audio_tsne} reveals that the embeddings of synthetic speech share a substantial distributional overlap with real clinical recordings. We observe two distinct primary clusters corresponding to the linguistic domains: an English cluster comprising ADReSS, ADReSSo, and SYN-EN, and a Mandarin cluster containing CIR-E and SYN-ZH. Crucially, within each linguistic domain, the synthetic samples exhibit substantial manifold alignment with their respective real-world clinical counterparts. We acknowledge that a slight distributional shift persists, which is likely attributable to the idealized acoustic environment of the synthetic generation compared to the background noise inherent in clinical recordings. Nevertheless, the extensive overlap indicates that our voice cloning pipeline successfully preserves the essential acoustic phenotypes and paralinguistic features required for robust, language-specific model training.

\subsection*{Diagnostic Performance and Cross-Linguistic Generalization}

To comprehensively evaluate the efficacy of the proposed framework, we conducted a rigorous comparative analysis against a diverse array of established and contemporary baselines. 
The evaluated open-source models include Qwen2-Audio-7B-Instruct~\cite{Chu2024Qwen2Audio}, MiniCPM-o-2.6~\cite{yao2024minicpm}, the Llama-3.1-based Ultravox series (v0.5 and v0.6), Phi-4-Multimodal-Instruct~\cite{abdin2024phi}, SeaLLMs-Audio-7B, R1-AQA~\cite{li2025reinforcement}, as well as the Qwen-Omni series comprising Qwen2.5-Omni-3B, Qwen2.5-Omni-7B~\cite{xu2025qwen2}, and Qwen3-Omni-30B-Instruct~\cite{xu2025qwen3}. Furthermore, to benchmark against the upper bound of current capabilities, we incorporated leading closed-source models, namely GPT-4o-Audio-Preview~\cite{hurst2024gpt}, Gemini-2.5-Flash~\cite{comanici2025gemini}, Gemini-3-Flash, and Gemini-3-Pro. 
A comprehensive summary of the model cards for these baselines is presented in Table~\ref{tab:model_names}.
Table~\ref{tab:main_results} summarizes the diagnostic performance across the three diverse cohorts.

\begin{table*}[h]
\centering
\fontsize{8}{11}\selectfont\setlength{\tabcolsep}{3pt}
\begin{threeparttable}
\caption{Performance comparison of different diagnostic models (\%)}
\label{tab:main_results}
\begin{tabular}{lcccccccccc}

\toprule
\multirow{2}{*}{\textbf{Method}}  
& \multicolumn{3}{c}{\textbf{EN: ADReSS}}   
& \multicolumn{3}{c}{\textbf{EN: ADReSSo}} 
& \multicolumn{3}{c}{\textbf{ZH: CIR-E}}
&  \textbf{AVG} \\

\cmidrule(lr){2-4} \cmidrule(lr){5-7} \cmidrule(lr){8-10} \cmidrule(lr){11-11}

& Macro-F1
& AVS 
& Bo8
& Macro-F1
& AVS
& Bo8
& Macro-F1
& AVS
& Bo8
& Macro-F1 \\

\midrule 
Qwen2-Audio-7B-Instruct & 49.35 (4.38) & 54.69 (4.13) & 58.33 & 46.62 (7.63) & 53.52 (4.88) & 60.56 & 30.07 (3.96) & 46.98 (2.93) & \textbf{52.94} & 42.01 (5.32) \\

MiniCPM-o-2.6 & 40.66 (5.29) & 51.82 (3.37) & 58.33 & 41.34 (4.67) & 51.41 (2.99) & 56.34 & 28.56 (2.19) & 45.67 (2.68) & 50.98 & 36.85 (4.05) \\

Ultravox-v0.5-Llama-3.1-8b & 46.87 (8.91) & 55.73 (5.39) & 66.67 & 43.75 (7.97) & 52.99 (5.12) & 63.38 & 26.07 (3.43) & 45.92 (1.90) & 49.02 & 38.90 (6.77) \\

Phi-4-Multimodal-Instruct & 55.43 (5.08) & 56.51 (4.47) & 62.50 & 48.91 (4.11) & 49.30 (3.98) & 53.52 & 24.73 (1.60) & 46.57 (0.91) & 48.37 & 43.02 (3.60) \\

Qwen2.5-Omni-3B & 60.05 (6.38) & 60.94 (6.48) & 68.75 & 58.09 (3.52) & 59.33 (3.62) & 64.79 & 28.25 (2.11) & 40.52 (3.17) & 44.44 & 48.80 (4.00) \\

Qwen2.5-Omni-7B & 57.50 (8.39) & 61.72 (5.60) & 72.92 & 59.43 (7.98) & 61.97 (5.32) & 73.24 & 31.39 (2.18) & 42.97 (2.79) & 49.02 & 49.44 (6.18) \\

SeaLLMs-Audio-7B & 56.76 (8.27) & 58.85 (6.97) & 70.83 & 50.79 (6.80) & 53.17 (5.40) & 61.97 & 28.45 (2.43) & 33.42 (2.99) & 39.22 & 45.33 (5.83) \\

R1-AQA & 49.77 (5.73) & 56.25 (3.13) & 62.50 & 50.18 (3.93) & 56.34 (2.82) & 61.97 & 27.51 (3.53) & 45.10 (3.50) & 49.67 & 42.49 (4.40) \\

Ultravox-v0.6-Llama-3.1-8b & 43.20 (6.12) & 52.08 (3.15) & 58.33 & 42.26 (4.52) & 50.70 (2.34) & 53.52 & 24.29 (2.20) & 46.00 (1.70) & 48.37 & 36.58 (4.28) \\

Qwen3-Omni-30B-Instruct & 74.35 (2.69) & 74.40 (2.66) & \underline{77.08} & 58.83 (1.88) & 58.98 (1.92) & 61.97 & 33.12 (2.64) & 44.12 (2.19) & 46.41 & 55.44 (2.41) \\

GPT-4o-Audio-Preview \dag & 51.60 (2.98) & 58.85 (2.02) & 62.50 & 50.41 (2.12) & 59.15 (1.22) & 61.97 & 26.17 (1.33) & 33.25 (1.65) & 35.29 & 42.73 (2.14) \\

Gemini-2.5-Flash \dag & 63.39 (5.92) & 63.54 (5.98) & 75.00 & 67.39 (2.93) & 67.43 (2.94) & 71.83 & 32.96 (3.33) & \underline{46.98} (1.89) & 50.98 & 54.58 (4.06) \\

Gemini-3-Flash \dag & 73.50 (2.82) & 73.96 (2.76) & 77.08 & 69.82 (3.45) & 70.77 (3.28) & 76.06 & 24.09 (2.27) & 29.00 (2.24) & 32.03 & 55.80 (2.89) \\

Gemini-3-Pro \dag & \underline{75.79} (4.06) & \underline{76.04} (4.03) & \textbf{83.33} & \underline{72.17} (4.01) & \underline{72.71} (3.85) & \textbf{80.28} & \underline{44.72} (1.90) & 44.61 (2.14) & 47.71 & \underline{64.23} (3.32) \\

\textbf{SynCog (Ours)} & \textbf{77.05} (2.95) & \textbf{77.34} (2.84) & \textbf{83.33} & \textbf{73.04} (2.29) & \textbf{73.59} (1.96) & \underline{77.46} & \textbf{48.71} (1.55) & \textbf{49.51} (1.38) & \underline{51.63} & \textbf{66.27} (2.34) \\

\bottomrule
\end{tabular}

\begin{tablenotes}
        \footnotesize
        \item[1] $\dag$ denotes closed-source models.
        \item[2] Metrics are reported as mean (sd) over eight runs with different random seeds.
        \item[3] The best results are bolded, and the second-best are underlined.
    \end{tablenotes}
\end{threeparttable}
\end{table*}

\subsubsection*{Performance on English Benchmarks}

To evaluate the core efficacy of SynCog in a well-studied linguistic context, we first benchmarked its performance on the standardized English ADReSS and ADReSSo datasets. When trained exclusively on synthetic data, SynCog achieved F1-scores of 77.05~($\pm$ 2.95)\% and 73.04~($\pm$ 2.84)\% on these benchmarks, respectively, as detailed in Table~\ref{tab:main_results}. 
Furthermore, SynCog exhibited superior diagnostic efficacy and potential, recording AVS scores of 77.34~($\pm$ 2.84)\% and 73.59~($\pm$ 1.96)\%, alongside a peak Bo8 performance of 83.33\% and 77.46\% on the two benchmarks. 
Notably, these zero-real-data results surpass the strongest closed-source baseline, Gemini-3-Pro, by margins of 1.26\% and 0.87\%. This achievement directly validates the capacity of our data synthesis pipeline to construct robust diagnostic models for English-speaking populations without relying on any real-world training samples.

\subsubsection*{Performance on Mandarin Real-world Cohorts}

Detecting fine-grained cognitive decline is inherently challenging due to the subtle, often imperceptible nature of early symptoms. Nevertheless, experimental results on the CIR-E cohort substantiate the robustness of our approach (Table~\ref{tab:main_results}). Specifically, SynCog attained superior diagnostic efficacy, securing an overall Macro-F1 score of 48.71~($\pm$ 1.55)\% and an AVS of 49.51~($\pm$ 1.38)\%. These results outperform leading closed-source models, including Gemini-3-Pro which recorded a Macro-F1 of 44.72~($\pm$ 1.90)\% and an AVS of 44.61~($\pm$ 2.14)\%, and Gemini-2.5-Flash which reached a Macro-F1 of 32.96~($\pm$ 3.33)\% and an AVS of 46.98~($\pm$ 1.89)\%. While Qwen2-Audio-7B-Instruct exhibited a higher Bo8 score of 52.94\%, SynCog delivered a more balanced and consistent diagnostic performance across all evaluated metrics, validating the cross-linguistic efficacy of the pipeline.

\begin{table}[h]
\centering
\fontsize{9}{11}\selectfont\setlength{\tabcolsep}{3pt}
\begin{threeparttable}
\caption{Stratified diagnostic performance on the CIR-E cohort utilizing SYN-ZH (\%)}
\label{tab:stratified_performance}
\begin{tabular}{llccc}
\toprule
\textbf{Diagnostic Objective} & \textbf{Clinical Transition} & \textbf{Sensitivity} & \textbf{Specificity} & \textbf{F1-score} \\
\midrule
Early Detection & MCI vs. HC & 40.03 (5.19) & 76.52 (5.19) & 53.02 (4.73) \\
Diagnostic Verification & AD vs. HC & 67.12 (4.41) & 69.70 (7.73) & 71.07 (3.74) \\
Clinical Screening & (MCI + AD) vs. HC & 83.54 (5.80) & 46.21 (9.91) & 84.15 (2.52) \\
\bottomrule
\end{tabular}
\begin{tablenotes}
            \item[1] Metrics are reported as mean (sd) over eight runs with different random seeds.
\end{tablenotes}
\end{threeparttable}
\end{table}

To address the clinical imperative for early identification, we stratified the diagnostic performance across specific clinical transitions, as summarized in Table~\ref{tab:stratified_performance}.
SynCog demonstrated a robust capability in differentiating early-stage impairment, achieving an F1-score of 53.02~($\pm$ 4.73)\% in the discrimination of MCI from HC. 
While detecting prodromal markers is challenging, the model maintained a high specificity of 76.52~($\pm$ 5.19)\%, ensuring reliable exclusion of healthy controls. 
This finding is particularly significant given the heterogeneous linguistic markers characterizing the prodromal phase of dementia. In the more pronounced AD vs. HC transition, where pathological speech markers are typically more pervasive, the model delivered a strong F1-score of 71.07~($\pm$ 3.74)\% with balanced sensitivity and specificity.
Most notably, in a simulated clinical screening scenario, defined as the binary classification of cognitively impaired individuals (comprising MCI and AD) against healthy controls, SynCog achieved a high F1-score of 84.15~($\pm$ 2.52)\%. 
This performance was underpinned by a high sensitivity of 83.54~($\pm$ 5.80)\%, demonstrating the model's efficacy as a potent screening tool capable of capturing the vast majority of cognitively impaired cases.
These stratified results underscore the diagnostic robustness of SynCog across the entire disease continuum. By explicitly modeling diverse demographic and cognitive profiles during synthesis, SynCog successfully bridges the linguistic gap, demonstrating that specialized diagnostic models for under-represented languages can be effectively constructed without relying on scarce real-world annotations.

\subsection*{Impact of Synthetic Data Scale}

To explicitly quantify the contribution of the synthetic data generation pipeline, an ablation study was conducted focusing on two aspects: the scaling ratio of synthetic augmentation and the impact of integrating different data sources.

\begin{figure}[h]
  \centering
  \includegraphics[width=0.65\linewidth]{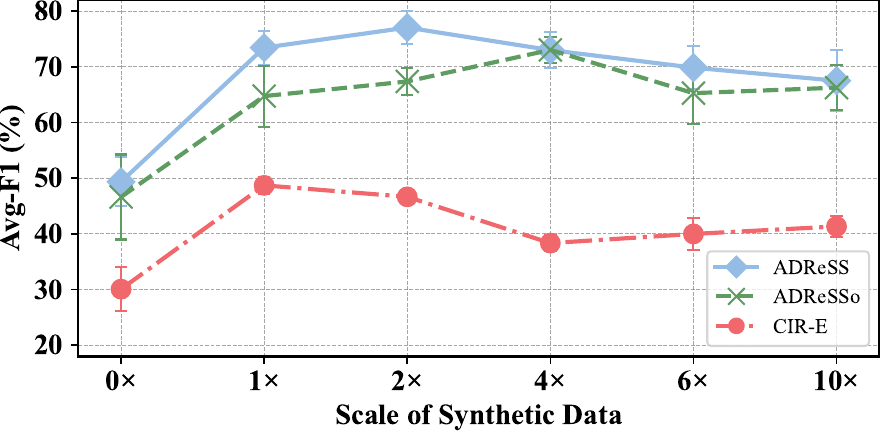}
  \caption{\textbf{Impact of data augmentation scaling on diagnostic performance.} The line graph illustrates the Average F1 score trajectories for the ADReSS, ADReSSo, and CIR-E datasets as the augmentation ratio scales from zero to five times the baseline. The error bars represent the standard deviation across experimental runs. The observed trends demonstrate a rapid initial performance improvement followed by a plateau or slight decline, indicating the existence of an optimal threshold for synthetic data integration.}
  \label{fig:syn_rate}
\end{figure}

Fig.~\ref{fig:syn_rate} illustrates the Average F1-score trajectories across the three datasets under increasing synthetic data scales. We observe a consistent pattern of rapid initial improvement followed by diminishing returns.
Specifically, expanding the synthetic data from the baseline (1$\times$) to a moderate scale (2$\times$) leads to a substantial performance gain, validating the effectiveness of data augmentation. However, a saturation point becomes apparent as the data scale continues to increase; beyond moderate expansion levels, additional synthetic samples result in marginal improvements or even slight performance degradation.
These results suggest that while synthetic data is beneficial for model adaptation, identifying an appropriate data scale is crucial to balance enhanced feature diversity against potential distribution shifts introduced by the generative process.

The ablation study further demonstrates that leveraging synthetic phenotypes to augment existing real-world English samples yields significant synergistic improvements across all metrics. Table~\ref{tab:ablation_results} presents the performance comparison across different data source configurations. Notably, models trained exclusively on synthetic samples (w/ Syn Data) consistently outperformed those trained on limited real-world datasets (w/ Real Data), with F1-scores improving from 75.96 ($\pm$ 3.51)\% and 72.62 ($\pm$ 2.46)\% to 77.05 ($\pm$ 2.95)\% and 73.04 ($\pm$ 2.29)\% on ADReSS and ADReSSo, respectively. This performance gain highlights the superior quality and diagnostic relevance of the generated synthetic phenotypes.  
Building upon this foundation, the integration of mixed data (w/ Mix Data), which was constructed by combining the original ADReSS and ADReSSo training sets with an optimized proportion of synthetic samples, further elevated diagnostic performance to peak levels, achieving F1-scores of 80.67~($\pm$ 3.26)\% and 78.46~($\pm$ 2.56)\%. Simultaneously, this synergistic approach drove substantial gains in diagnostic accuracy, with AVS rising to 80.73\% and 78.52\%, while the performance ceiling was significantly expanded, as evidenced by the elevated Bo8 scores of 87.50\% and 83.10\%. 
These findings demonstrate that synthetic data serves not only as a high-fidelity alternative but also as a catalyst that significantly enhances diagnostic precision in data-scarce clinical settings.
This finding suggests that real and synthetic data provide complementary supervision signals. The real data anchors the model to the ground-truth clinical distribution, while the diverse synthetic samples cover the long-tail pathological features and decision boundaries that are sparsely represented in limited clinical datasets.

\begin{table}[h]
\centering
\fontsize{9}{11}\selectfont\setlength{\tabcolsep}{3pt}
\begin{threeparttable}
\caption{Ablation study on data source configurations for the SynCog (\%)}
\label{tab:ablation_results}
\begin{tabular}{lcccccc}

\toprule
\multirow{2}{*}{\textbf{Method}}  
& \multicolumn{3}{c}{\textbf{EN: ADReSS}}   
& \multicolumn{3}{c}{\textbf{EN: ADReSSo}}  \\

\cmidrule(lr){2-4} \cmidrule(lr){5-7} 

& Macro-F1
& AVS
& Bo8
& Macro-F1
& AVS
& Bo8 \\

\midrule 
Zero-Shot &  49.35 (4.38) & 54.69 (4.13) & 58.33 & 46.62 (7.63) & 53.52 (4.88) & 60.56 \\

$\Delta$  & +26.61 & +21.35 & +22.92 & +26.00 & +19.37 & +15.50 \\

w/ Real Data & 75.96 (3.51) & 76.04 (3.45) & 81.25 & 72.62 (2.46) & 72.89 (2.41) & 76.06 \\

$\Delta$ & +1.09 & +1.30 & +2.08 & +0.42 & +0.70 & +1.40 \\

w/ Syn Data & 77.05 (2.95) & 77.34 (2.84) & 83.33 & 73.04 (2.29) & 73.59 (1.96) & 77.46 \\

$\Delta$& +3.62 & +3.39 & +4.17 & +5.42 & +4.93 & +5.64 \\

w/ Mix Data & 80.67 (3.26) & 80.73 (3.25) & 87.50 & 78.46 (2.56) & 78.52 (2.51) & 83.10 \\

\bottomrule
\end{tabular}
\begin{tablenotes}
            \item[1] Metrics are reported as mean (sd) over eight runs with different random seeds.
\end{tablenotes}
\end{threeparttable}
\end{table}

\section*{Discussion}
\label{sec:discussion}

This study introduces SynCog, a novel framework that bridges the gap between generative AI and clinical utility by integrating persona-driven phenotypic simulation with reasoning self-distillation. Diverging from prior studies constrained by surface-level data augmentation, our approach constructs clinically high-fidelity digital subjects anchored in established pathological profiles. This innovation enables the training of robust diagnostic models that exhibit superior diagnostic efficacy, outperforming leading closed-source baselines such as Gemini-3-Pro, while effectively mitigating the privacy risks and scarcity bottlenecks inherent in sensitive clinical data. By demonstrating that generative simulation can serve as a primary source of reliable supervision, SynCog offers a transformative perspective on digital phenotyping and challenges the conventional dependency on large-scale, manually annotated clinical datasets.

The superior performance of our framework stems from the synergistic supervision provided by the integration of real and synthetic data. As evidenced by our ablation studies, while synthetic data alone enables robust generalization, the hybrid approach yields the most accurate diagnostic boundaries. We attribute this to the complementary roles of the two data sources: real data serves as a distributional anchor to stabilize the model against artifacts, while diverse synthetic samples densify the feature space. This densification covers long-tail pathological patterns, such as specific syntactic deficits or subtle paralinguistic cues, which are often sparsely represented in limited clinical cohorts. Furthermore, the distinct overlap of acoustic feature distributions confirms that our voice cloning pipeline successfully preserves essential disease-specific phenotypes. However, we observed a performance plateau and subsequent decline with excessive synthetic augmentation. This suggests that data balance is critical; an optimal ratio is required to maximize feature diversity while minimizing the risk of introducing synthetic artifacts that could skew the clinical distribution.

Beyond diagnostic accuracy, the clinical utility of an AI tool relies heavily on trust and transparency. Traditional deep learning models often function as opaque systems that provide probability scores without actionable rationale. Through reasoning self-distillation, SynCog transforms this paradigm by explicating the diagnostic pathway. By learning to articulate specific linguistic evidence, such as empty speech or circumlocution, consistent with the final prediction, our model mimics the cognitive process of a clinician. This interpretability is not merely a technical feature but a prerequisite for integration into clinical workflows, as it empowers physicians to validate AI-generated assessments and mitigates the risk of unsubstantiated or hallucinated diagnoses.

Current digital health research is predominantly centered on English-speaking populations, exacerbating global health disparities. The robust performance of SynCog on the Mandarin CIR-E cohort highlights a scalable pathway to bridge this linguistic gap. Unlike traditional methods that require collecting expensive clinical datasets for every target language, our framework facilitates rapid adaptation through persona-based synthesis. By simply adjusting the demographic and linguistic parameters of the generative prompt, we can construct specialized diagnostic models for under-represented languages or dialects. This capability is instrumental in fostering global health equity, offering a cost-effective solution to deploy standardized cognitive screening tools in low-resource settings.

The non-invasive and cost-effective nature of SynCog positions it as an ideal candidate for large-scale community screening and telemedicine. As a digital biomarker, speech can be collected remotely via ubiquitous mobile devices, significantly lowering accessibility barriers compared to invasive biomarkers or neuroimaging. In a tiered healthcare system, our framework could serve as an accessible gatekeeper to identify individuals with early cognitive decline who warrant further specialized examination, such as positron emission tomography (PET) scans or cerebrospinal fluid (CSF) analysis. This proactive screening capability is particularly vital for aging societies, enabling timely intervention and more efficient clinical resource allocation.

Despite the promising results, our study has several limitations that warrant consideration.

One significant limitation lies in the potential biological disconnect within the acoustic synthesis pipeline. Neurodegenerative disorders entail not only linguistic simplification but also profound deficits in neuromuscular control. These impairments often manifest as micro-tremors, irregular articulatory breakdowns, and hesitancy induced by cognitive load. However, our current two-stage framework generates narratives via large language models followed by a separate acoustic synthesis step. This separation may inadvertently decouple linguistic content from these biological constraints. While the employed voice cloning technique effectively captures the static timbre and general voice quality of elderly speakers, it may not fully replicate the dynamic pathological prosody stemming from actual brain lesions. Consequently, the synthetic speech might lack specific fine-grained motor speech markers and potentially creates a domain gap when the model is applied to real-world patients exhibiting complex neuro-acoustic symptoms.

Furthermore, the scope of our evaluation is currently confined to the picture description paradigm. While this task significantly simplifies the assessment workflow, relying on a single neuropsychological probe restricts the holistic evaluation of cognitive function. A comprehensive diagnosis typically encompasses multiple cognitive domains, including verbal fluency, episodic memory recall, and executive function tasks. Future iterations of SynCog will aim to extend the generative framework to encompass a broader battery of cognitive tests and thereby enhance the ecological validity of the screening tool.

A final challenge pertains to clinical safety and the risk of reasoning hallucination. Although CoT fine-tuning significantly improves interpretability, there remains a possibility that the model could generate a plausible-sounding medical rationale that does not accurately reflect the internal features driving its prediction. This phenomenon is known as the unfaithfulness of explanation and poses a challenge for clinical integration. Future work will focus on integrating mechanistic interpretability techniques to ensure that the generated natural language explanations are causally aligned with the internal decision-making processes of the model.

In conclusion, SynCog provides a unified, privacy-preserving, and interpretable framework for AI-driven cognitive assessment. By harmonizing generative simulation with clinical reasoning, we demonstrate that high-performance diagnostic tools can be built with minimal reliance on real-world data. This work lays the foundation for the next generation of ethical and scalable digital health solutions, moving us closer to the goal of accessible cognitive healthcare for all.

\section*{Methods}
\label{sec:methods}

\subsection*{Diagnostic Task Formulation}

We formulate the cognitive assessment of picture description tasks as a supervised classification problem.
Given $\mathcal{D} = \{(x_i, y_i)\}_{i=1}^M$, each sample $x_i = (a_i, t_i)$ contains an audio recording $a_i$ and its transcript $t_i$, with label $y_i$ indicating the cognitive status.
Our primary objective is to learn a mapping function $f(x_i) \to \hat{y}_i$ to assess cognitive status.

To adapt this classification task to MLLMs, we recast it as a structured text generation problem. Given a prompt template $P_{cls}$ that embeds the sample $x_i$, the MLLM generates a response $S_i = \mathcal{LLM}(P_{\textit{cls}}(x_i))$. The output $S_i$ is constrained to a predefined format, from which the predicted label $\hat{y}_i$ can be parsed.

\subsection*{Data Synthesis}

\subsubsection*{Persona-Based Digital Phenotyping}

To account for inter-subject variability in cognitive assessments, we construct a controllable persona model that conditions the generative process. Each persona is defined by demographic attributes, including gender, age, and education level, together with the cognitive status label. These factors are motivated by clinical evidence indicating that demographic and educational characteristics substantially influence linguistic and cognitive performance~\cite{long2020understudied, kim2017factors, bennett2003education, podcasy2016considering}.

Beyond demographics, each persona is characterized by a five-dimensional linguistic style vector $v = (v_1, v_2, v_3, v_4, v_5)$, encompassing narrative length, syntactic complexity, spatial reference, fluency, and clarity. 
Specifically, $v$ is formulated as a quantized style embedding that encapsulates cognitive-functional signatures into a latent representation.
To enable granular control, each dimension is discretized into three ordinal levels: \textit{poor}, \textit{normal}, and \textit{good}. The style parameters are sampled from truncated Gaussian distributions, where the latent means are anchored to the cognitive status and further modulated by demographic factors.
Formally, the continuous latent style $\tilde v_k$ is sampled as:
\begin{equation}
\tilde v_k \sim \mathcal{N}\Big(
    \mu_k^{(y)} - \alpha_k\, g(\text{age}) + \beta_k\, h(\text{edu}),\ \sigma_k^2
\Big),
\end{equation}
The final discrete style level $v_k$ is obtained through clipping and rounding operations:
\begin{equation}
v_k = \operatorname{Round}(\operatorname{Clip}(\tilde v_k, 1, 3)),
\end{equation}
where $\tilde v_k$ denotes the continuous latent variable, $\mu_k^{(y)}$ represents the baseline mean for cognitive status $y$, and $\alpha_k$ and $\beta_k$ are sensitivity coefficients controlling the influence of age and education respectively. The functions $g(\text{age})$ and $h(\text{edu})$ represent the mappings of age and education level, while $\sigma_k^2$ is the variance capturing natural variability. The $\operatorname{Clip}(\cdot)$ function constrains the value to the range $[1,3]$, and $\operatorname{Round}(\cdot)$ quantizes the continuous value into a discrete integer in $\{1,2,3\}$, corresponding to the three ordinal levels.

The standardized scoring rubrics governing each linguistic dimension are detailed in Table~\ref{tab:text_criteria}. To ensure that the synthetic samples reflect realistic variability across subjects while remaining consistent with established clinical observations, the resulting persona specification is embedded into a structured prompt template for LLM generation (See Supplementary Fig.~\ref{fig:personal_template} for prompt details).


\subsubsection*{Text Generation}

The synthesized persona specification is integrated into a structured prompt to guide the LLM-based text generation, ensuring that the outputs capture the demographic and cognitive heterogeneity inherent in clinical populations.
In this framework, the advanced MLLM GPT-4o is utilized to generate picture description narratives conditioned jointly on the visual stimulus $I$ and the constructed persona $p_i$. 

Formally, given an image $I$ and its associated persona specification $p_i$, 
the generation process can be formulated as
\begin{equation}
T_i = \mathcal{LLM}_{\textit{gpt}}\big(P_{\textit{syn}}(I, p_i)\big),
\end{equation}
where $P_{syn}$ denotes the prompt construction function that embeds demographic attributes and linguistic style parameters into a textual instruction, $\mathcal{LLM}_{\textit{gpt}}(\cdot)$ represents the GPT-4o, and $T_i$ is the generated picture description.

To ensure clinical fidelity, the prompt design explicitly incorporates the five-dimensional linguistic style vector to guide the model toward emulating colloquial oral discourse rather than formal, structured summaries. This formulation enables controllable text generation that accurately reflects the spontaneous linguistic variability and pathological markers observed in real-world subjects. 

\subsubsection*{Speech Synthesis}

To transform the generated narratives into acoustic signals, a specialized geriatric timbre library was curated specifically to preserve the acoustic integrity of elderly speech. 
This library was constructed by selecting reference audios from high-quality clinical corpora and public datasets~\cite{liang2025construction, chen2023raw}, ensuring an age-range match with the target persona to encapsulate representative geriatric vocal traits. 
To ensure clean and uniform audio quality, a rigorous preprocessing pipeline was implemented, which incorporated adaptive noise reduction, silence removal, and sampling-rate normalization. The resulting repository comprises 551 female and 436 male distinctive timbre profiles, capturing a wide spectrum of geriatric acoustic characteristics, such as increased jitter, shimmer, and reduced fundamental frequency, which are representative of the target demographic.

The objective of this stage is to ensure that the synthesized audio not only conveys the linguistic content $T_i$ but also adheres to the demographic constraints defined in the persona specification $p_i$, particularly regarding gender and age-specific vocal signatures. IndexTTS2~\cite{zhou2025indextts2} is employed for reference-based voice cloning. Given the generated text and a sex-matched reference audio sample $a_i^{\textit{ref}}$ from the library, the synthetic waveform $\hat{a}_i$ is generated as:

\begin{equation}
\hat{a}_i = \mathcal{T}_{\textit{Index}}(T_i, a_i^{\textit{ref}}),
\end{equation}
Here $\mathcal{T}(\cdot)$ denotes the IndexTTS2 inference function, which performs reference-based voice cloning: it replicates the speaker’s timbre from the reference audio $a_i^{\textit{ref}}$ while preserving the prosody and linguistic content of $T_i$.

\begin{table*}[t]
\centering
\fontsize{8}{11}\selectfont\setlength{\tabcolsep}{3pt}
\caption{Assessment dimensions and scoring criteria}
\begin{tabular}{c
                >{\arraybackslash}p{2.8cm}
                >{\arraybackslash}p{2.8cm}
                >{\arraybackslash}p{2.8cm}
                >{\arraybackslash}p{2.8cm}
                >{\arraybackslash}p{2.8cm}}

\toprule
\textbf{Score} & 
\textbf{Narrative Length} & 
\textbf{Syntactic Complexity} & 
\textbf{Spatial Expressions} & 
\textbf{Speech Fluency} & 
\textbf{Clarity of Expression} \\

\midrule
1 & 60--110 words; only 2--3 key elements; minimal descriptive detail. 
  & Predominantly short, simple sentences; rare conjunctions; absence of subordinate clauses. 
  & 0--1 spatial reference; vague or imprecise expression. 
  & $\geq$4 fillers or word-search events; frequent pauses/elongations; disrupted speech flow. 
  & $\geq$4 vague terms; $\leq$1 key noun; frequent pronoun reliance. \\
\midrule
2 & 90--150 words; covers main content with 3--4 descriptive details. 
  & Mix of short and medium-length sentences; occasional conjunctions; may contain simple subordinate clauses. 
  & 1--2 explicit spatial references; basic relative positioning expressed. 
  & 2--3 disfluency events; occasional pauses or self-repairs; overall understandable. 
  & 2--3 vague terms; 2--3 key nouns; occasional pronoun substitution. \\
\midrule
3 & 120--200 words; relatively complete description with $\geq$4 details; well-organized structure. 
  & Alternation of short and long sentences; inclusion of causal, temporal, or conditional relations; coherent cohesion. 
  & $\geq$3 explicit spatial references; comprehensive expression of spatial relations. 
  & $\leq$1 disfluency event; natural speech rate; minimal pauses or word-searching. 
  & $\leq$1 vague term; $\geq$3 key nouns; predominantly specific references. \\
\bottomrule
\end{tabular}
\label{tab:text_criteria}
\end{table*}

\subsection*{CoT deduction Fine-Tuning}

Direct diagnostic label prediction using MLLMs often exhibits instability and limited generalization, as models may gravitate toward superficial linguistic cues rather than cognitively substantive markers. To mitigate this risk of shortcut learning, we implement a reasoning distillation framework that integrates CoT paradigm with SFT (see Supplementary Fig.~\ref{fig:cot_template} for details). This approach guides the model to develop interpretable and robust diagnostic pathways by mimicking clinical deliberative processes.

Within this framework, the model $\mathcal{M}$ is tasked with generating explicit diagnostic rationales that bridge multimodal evidence comprising both linguistic and acoustic features to the final assessment. Given a multimodal input $(a_i, t_i)$ and its corresponding ground-truth label $y_i$, the model is prompted to infer a reasoning sequence $r_i$ that elucidates the underlying decision logic:
\begin{equation}
r_i = \mathcal{LLM}\big(P_{\textit{cot}}(a_i, t_i, y_i)\big).
\end{equation}

The generated reasoning traces $r_i$ are then paired with their corresponding inputs and labels to construct a reasoning-augmented dataset:
\begin{equation}
\mathcal{D}_{\textit{self-cot}} = \{(a_i, t_i, r_i, y_i)\}_{i=1}^M.
\end{equation}

This dataset serves as structured supervision, enabling the model to learn a mapping from complex multimodal features to clinical rationales. By fine-tuning on $\mathcal{D}_{\textit{self-cot}}$, the foundational model transitions into the specialized diagnostic model $\mathcal{M}$, evolving from a black-box predictor into a transparent tool capable of articulating an evidentiary basis. The comprehensive training pipeline is detailed in Algorithm~\ref{alg:algorithm}.

\begin{algorithm}[t]
\caption{Pipeline of the SynCog Framework}
\label{alg:algorithm}
\begin{algorithmic}[1] 
\Require 
    Dataset $\mathcal{D}=\{((a_i, t_i), y_i)\}_{i=1}^{N}$; image stimuli $\{I\}, \mathcal{V}$; LLM $\mathcal{M}$; prompts $\{P_{\text{syn}}, P_{\text{cls}}, P_{\text{cot}}\}$
\Ensure
Fine-tuned model $\mathcal{M}$; predictions $\hat{y}$
\State \textbf{Step 1: Persona-Based Data Synthesis}
\For{$i=1$ \textbf{to} $N_{\operatorname{syn}}$}
    \State Sample demographics \& style $p_i$ conditioned on target $y_i$
    \State $T_i \gets \mathcal{LLM}_{\operatorname{gpt}}(P_{\operatorname{syn}}(I_i,p_i))$
    \State Select $r_i^{\operatorname{ref}} \in \mathcal{V}$ (sex-matched)
    \State $\hat{a}_i \gets \mathcal{T}_{\operatorname{Index}}(T_i, r_i^{\operatorname{ref}})$
    \State $\mathcal{D}' \gets \mathcal{D} \cup \{((\hat{a}_i,T_i), y_i)\}$
\EndFor

\State \textbf{Step 2: CoT deduction Fine-Tuning}
\State Initialize $\mathcal{D}_{\text{self-cot}} \gets \emptyset$
\For{each $(x_i,y_i) \in \mathcal{D}'$}
    \State $r_i \gets \mathcal{M}(P_{\operatorname{cot}}(a_i,t_i,y_i))$
    \State $\mathcal{D}_{\operatorname{self-cot}} \gets \mathcal{D}' \cup \{(a_i,t_i,r_i,y_i)\}$
\EndFor
\State Fine-tune $\mathcal{M}$ on $\mathcal{D}_{\text{self-cot}}$ with SFT

\State \textbf{Step 3: Inference for Cognitive Impairment Assessment}
\For{each test sample $x_j=(a_j,t_j)$}
    \State $S_j \gets \mathcal{M}(P_{\operatorname{cls}}(x_j))$
    \State Parse $\hat{y}_j$ from $S_j$
    \State \Return $\hat{y}$
\EndFor
\end{algorithmic}
\end{algorithm}

\subsection*{Diagnostic Inference}

During the inference phase, the fine-tuned model $\mathcal{M}$ is presented with multimodal inputs consisting of an audio recording $a_j$ and its corresponding transcript $t_j$. Rather than producing a solitary diagnostic category, the model is prompted to first articulate a reasoning sequence $r_j$. This heuristic reasoning ensures that the final assessment is grounded in specific pathological markers, such as reduced syntactic complexity, frequent speech disfluency, or diminished clarity of expression. 
The integration of CoT deduction effectively mitigates the risk of shortcut learning by forcing the model to align multimodal evidence with the standardized scoring rubrics. By synthesizing evidence from both acoustic prosody and linguistic content, the model generates a structured response $S_j$, from which the final cognitive status $\hat{y}_j$ is parsed (Supplementary Fig.~\ref{fig:Inference_template}). This transparent inference pathway not only enhances the robustness of the diagnostic results across different linguistic contexts but also provides clinicians with interpretable evidence to support the final assessment.

\subsection*{Data Pre-processing}

A unified preprocessing pipeline was implemented to ensure consistent data quality and the preservation of diagnostic markers across all evaluation cohorts. To standardize the acoustic input, raw audio recordings were first converted to a mono-channel format and resampled to 16\,kHz. The participant's speech was isolated from the original dialogues using the \textit{speaker-diarization-3.1} pipeline~\cite{plaquet2023powerset}, which effectively removed interviewer prompts and instructions throughout the task. Since interviewer involvement was primarily limited to providing instructions, the extraction of participant-specific segments ensures the clinical integrity and completeness of the diagnostic content.

The isolated acoustic sequences were subsequently transcribed using the \textit{faster-whisper} model to generate verbatim transcripts. During this process, we applied standard punctuation and case normalization while explicitly retaining all disfluencies and filler words (e.g., ``uh'', ``um''). Preserving these hesitation markers is crucial for capturing cognitive deficits characterized by speech fragmentation, as they serve as vital linguistic biomarkers for neurodegenerative assessment. To accommodate the input constraints of LLMs, the total duration of each concatenated audio sample was capped at 90 seconds. Each processed sample was ultimately represented as a synchronized multimodal pair $(a_i, t_i)$, comprising the participant's full concatenated acoustic sequence and the corresponding verbatim transcript.

\subsection*{Evaluation Metrics}
To comprehensively evaluate the robustness and efficacy of the proposed method, each model is executed for $N$ independent stochastic rollouts. We employ three distinct metrics to quantify performance: the Macro F1-score (Macro-F1), the Average Score (AVS), and the Best-of-N performance (BoN).

First, we define the standard classification statistics for each sample: True Positives (TP), False Positives (FP), True Negatives (TN), and False Negatives (FN). Based on these, the per-class precision ($P_c$) and recall ($R_c$) are computed as:

\begin{equation}
P_c = \frac{TP_c}{TP_c + FP_c}, \quad 
R_c = \frac{TP_c}{TP_c + FN_c},
\end{equation}
and the per-sample accuracy is:

\begin{equation}
Acc_i = \frac{TP_i + TN_i}{TP_i + TN_i + FP_i + FN_i}.
\end{equation}

\subsubsection*{Macro F1-score (Macro-F1)}
To assess the classification performance of the model while accounting for potential class imbalance, we utilize the Macro F1-score. It is calculated as the arithmetic mean of the per-class F1 scores over all $N$ rollouts:

\begin{equation}
F1 = \frac{1}{N} \sum_{n=1}^{N} \left( \frac{1}{C} \sum_{c=1}^{C} \frac{2 \cdot P_c^{(n)} \cdot R_c^{(n)}}{P_c^{(n)} + R_c^{(n)}} \right),
\end{equation}
where $C$ is the number of classes, and $P_c^{(n)}$ and $R_c^{(n)}$ denote the precision and recall for class $c$ in rollout $n$.

\subsubsection*{Average Score (AVS)}
The Average Score serves as a measure of the expected performance and stability of the model across repeated trials. Let $Acc_i^{(n)}$ denote the accuracy of the $i$-th sample in rollout $n$, where $i \in \{1, \dots, M\}$ and $n \in \{1, \dots, N\}$. The AVS is computed as the mean accuracy over all samples and all $N$ rollouts:

\begin{equation}
AVS = \frac{1}{N} \sum_{n=1}^{N} \left( \frac{1}{M} \sum_{i=1}^{M} Acc_i^{(n)} \right),
\end{equation}
where $M$ is the total number of samples.

\subsubsection*{Best-of-N (BoN)}
To evaluate the peak potential of the generation capability of the model, we report the BoN metric. This metric captures the upper bound of performance by selecting the maximum accuracy achieved among the $N$ rollouts:

\begin{equation}
BoN = \max_{n=1,2,\dots,N} \left( \frac{1}{M} \sum_{i=1}^{M} Acc_i^{(n)} \right),
\end{equation}
where $Acc_i^{(n)}$ denotes the accuracy of the $i$-th sample in rollout $n$. 
BoN provides an estimate of the maximum achievable performance of the model.

\subsection*{Implementation Details}

For all LLM-based inferences, decoding parameters were standardized across models to ensure a fair comparison. Inference was conducted in \textit{bfloat16} precision to reduce memory consumption, with a rollout performed $N=8$ times.

During fine-tuning, we employed the AdamW optimizer. The model was trained for 5 epochs, with the learning rate gradually decayed following a cosine annealing schedule. The batch size per device was set to 1, and gradient accumulation steps were set to 8, resulting in an effective batch size of 8. The dropout rate was set to 0.1.
To determine the optimal hyperparameters, we conducted a grid search over the learning rate and the LoRA rank $r$. The learning rate candidates were 
$\{1\mathrm{e}{-3},\, 5\mathrm{e}{-4},\, 1\mathrm{e}{-4},\, 5\mathrm{e}{-5},\, 1\mathrm{e}{-5}\}$, 
and the candidate ranks were $r \in \{8,\, 16,\, 32\}$. The scaling factor $\alpha$ in LoRA was set to twice the selected rank $r$, and LoRA was applied to all attention projection layers, including the query, key, value, and output projections.
All experiments were implemented on 8 NVIDIA RTX 3090 GPUs.



\section*{Code availability}
The source code supporting this study is publicly available at
\url{https://github.com/FengRui1998/SynCog}.






\bibliography{reference}     
\label{MainEnd}

\clearpage 
\appendix

\makeatletter
  \let\OldThePage\thepage      
  \renewcommand{\thepage}{S\arabic{page}}
  \setcounter{page}{1}
\makeatother

\noindent
{\LARGE\bfseries Supplementary Information: \\Cross-Linguistic Persona-Driven Data Synthesis for Robust Multimodal Cognitive Decline Detection}     

\setcounter{figure}{0}
\renewcommand{\thefigure}{S\arabic{figure}}

\setcounter{table}{0}
\renewcommand{\thetable}{S\arabic{table}}

\rfoot{\small\sffamily\bfseries \thepage/\pageref{SupEnd}}

\section{Prompt Design}

In this section, we provide the detailed prompt templates employed in the SynCog framework. The prompt engineering strategy is divided into three distinct stages: persona-based data synthesis, reasoning extraction, and diagnostic inference.

\begin{figure}[h!]
  \centering
  \includegraphics[width=0.9\linewidth]{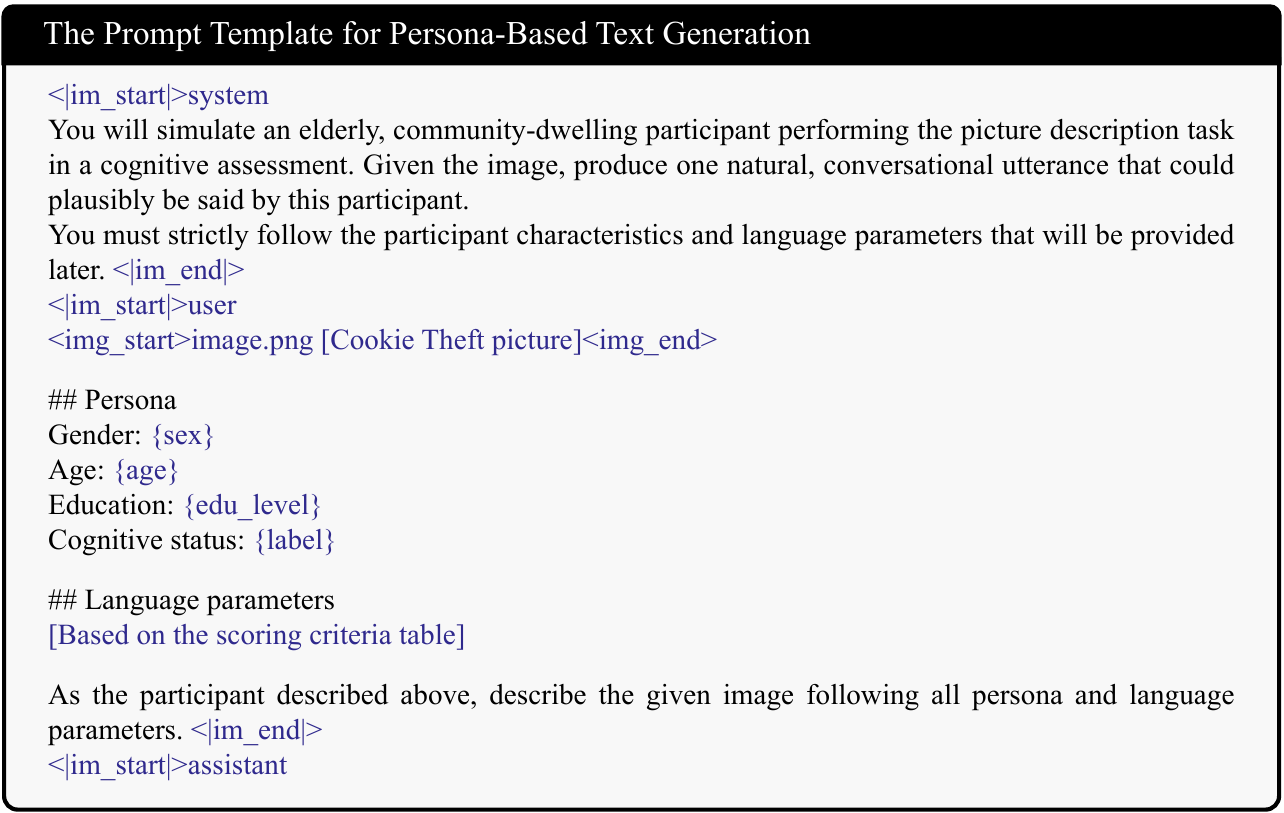}
  \caption{\textbf{Prompt Template for Persona-Based Text Generation.} The prompt conditions the text generation on specific demographic attributes and a discrete linguistic style vector. This mechanism enforces the production of narratives that reflect specific cognitive deficits rather than generic descriptions.}
  \label{fig:personal_template}
\end{figure}

\begin{figure}[h!]
  \centering
  \includegraphics[width=0.9\linewidth]{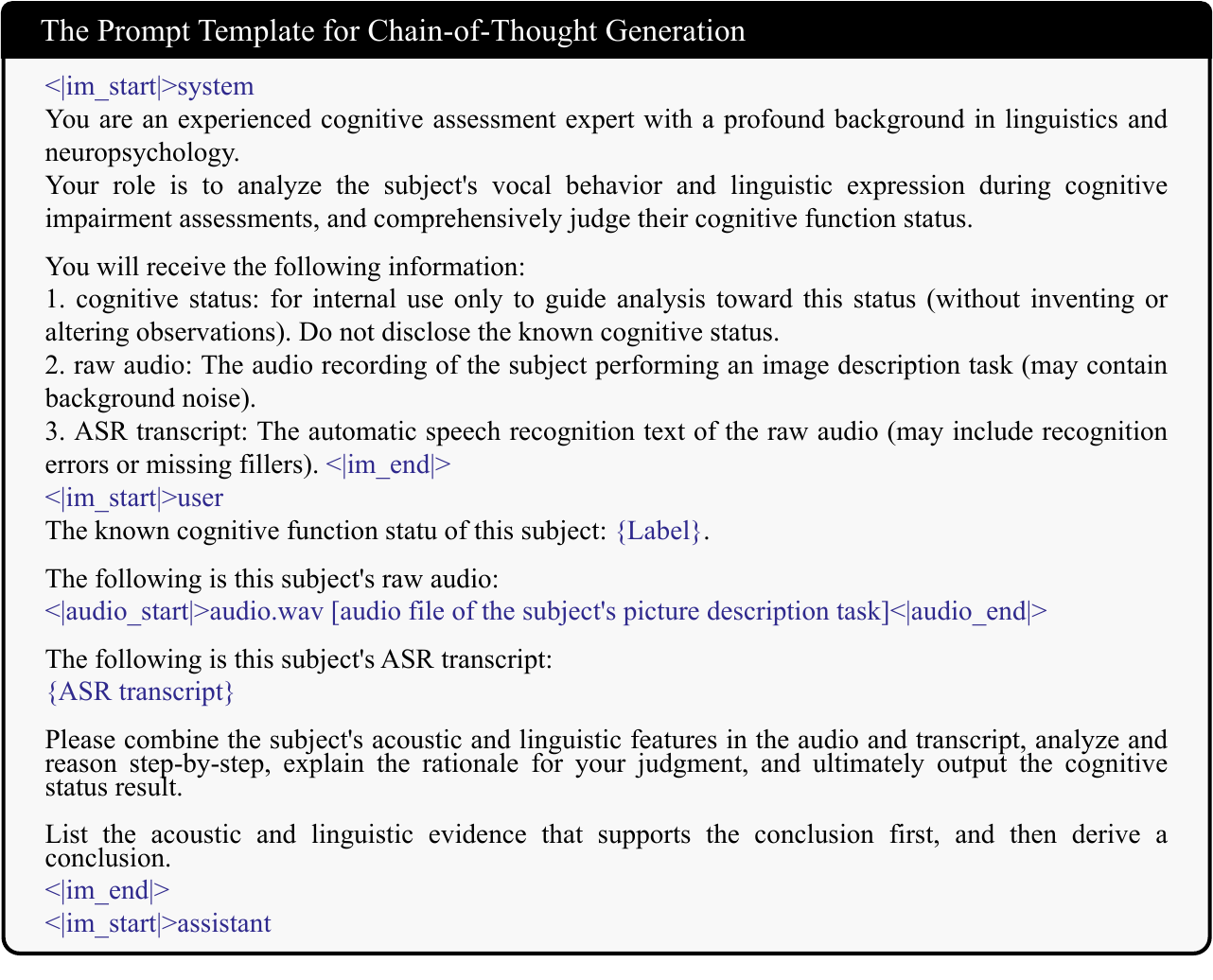}
  \caption{\textbf{Prompt template for label-conditioned reasoning generation.} This template is utilized during the training data construction phase. By incorporating the ground-truth diagnostic label, the prompt guides the model to perform hindsight analysis, explicitly articulating the linguistic and logical evidence that supports the correct diagnosis. The resulting output serves as the CoT supervision.}
  \label{fig:cot_template}
\end{figure}

\begin{figure}[h!]
  \centering
  \includegraphics[width=0.9\linewidth]{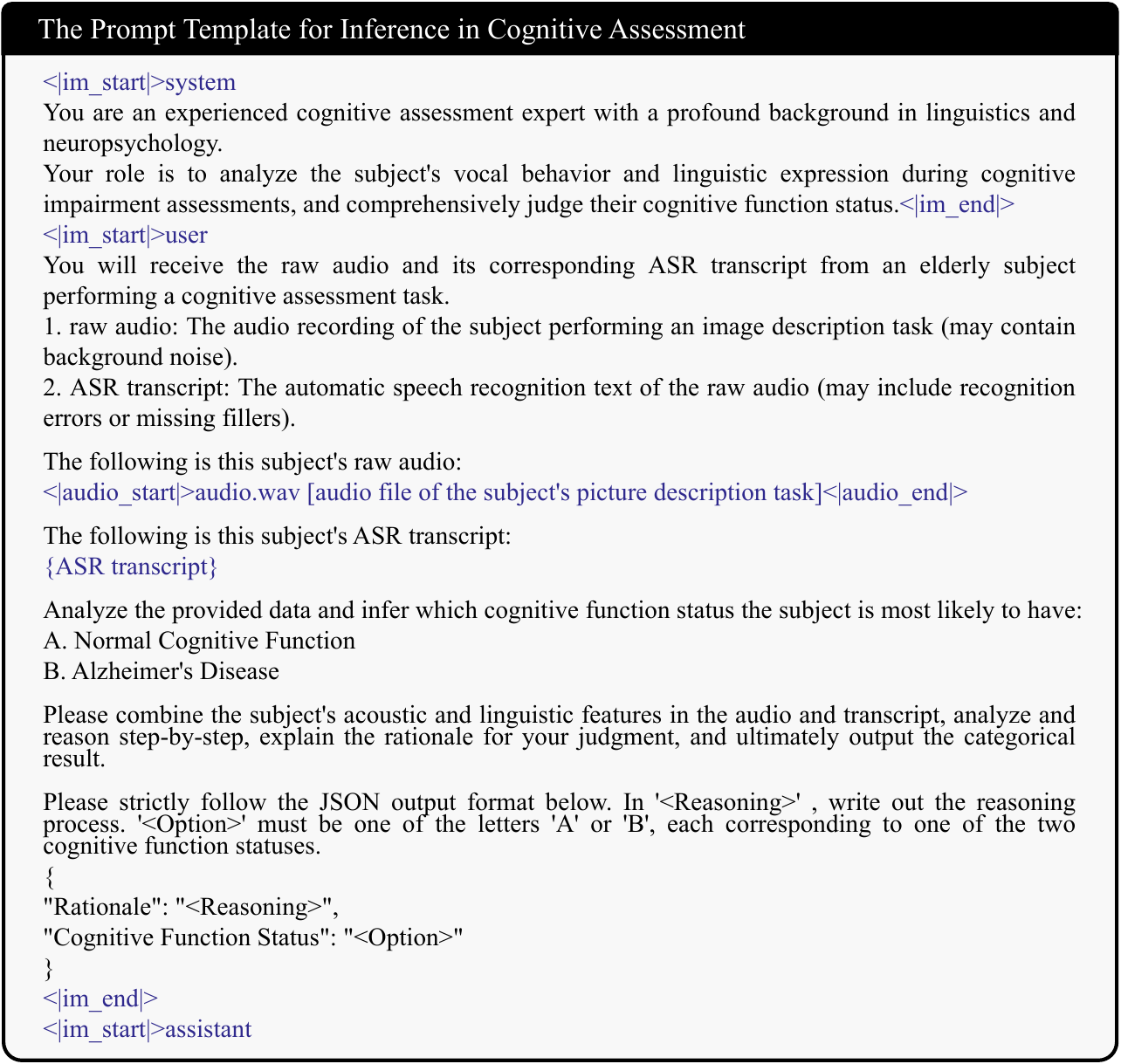}
  \caption{\textbf{Prompt template for multimodal diagnostic inference.} The LLM receives the task instruction along with the participant's audio and text records. The model is instructed to analyze linguistic patterns and generate a predicted cognitive status.}
  \label{fig:Inference_template}
\end{figure}

To generate diverse and clinically plausible synthetic phenotypes, we designed a persona-conditioned generation prompt ($P_{\textit{syn}}$), as illustrated in Fig.~\ref{fig:personal_template}. This template acts as a bridge between the structured persona specifications and natural language narratives. It explicitly integrates the sampled demographic attributes (e.g., age, education) and the five-dimensional linguistic style vector into the instruction. By conditioning the LLM on these fine-grained parameters, we ensure that the generated picture descriptions strictly adhere to the simulated cognitive profile rather than reverting to the model's default, high-quality writing style.

For the construction of the reasoning-augmented dataset, we utilized a label-conditioned prompt ($P_{\textit{cot}}$), shown in Fig.~\ref{fig:cot_template}. Unlike the inference stage, this template incorporates the ground-truth diagnosis to guide the model in a "hindsight analysis" process. The prompt instructs the model to reverse-engineer the diagnostic logic, explicitly articulating the specific linguistic and acoustic evidence that supports the known label. These high-quality, evidence-based reasoning traces are subsequently used as supervision targets for the CoT fine-tuning.

Finally, to facilitate the actual diagnostic task during the testing phase, we designed a zero-shot inference prompt ($P_{\textit{cls}}$), as illustrated in Fig.~\ref{fig:Inference_template}. This template instructs the LLM to act as a cognitive assessment specialist. It is provided with the multimodal input without the clinical label. The model is tasked with analyzing the linguistic patterns to predict the cognitive status and, thanks to the prior CoT fine-tuning, spontaneously generating the supporting rationale.

\section{Model Card}

To ensure benchmarking transparency and reproducibility, we provide specifications for the Multimodal Large Language Models (MLLMs) evaluated in this study (Table~\ref{tab:model_names}). The selection encompasses a representative spectrum of architectures:

\begin{itemize}[leftmargin=10px]
\item \textbf{Open-source Models:} Includes the Qwen, Ultravox, Phi-4, and MiniCPM series, ranging from 3B to 30B parameters. These models facilitate assessments of diagnostic scaling and cross-architectural generalizability. 

\item \textbf{Closed-source Models:} Includes GPT and the Gemini series. These models represent current leading benchmarks in multimodal inference. Due to their commercial nature, their internal architectural parameters are undisclosed, and they are accessed exclusively via API interfaces.
\end{itemize}

\begin{table*}[t]
\centering
\small
\caption{Model cards for MLLMs}
\begin{tabular}{l|c|l}
\toprule
\textbf{Model} & \textbf{\# Params} & \textbf{Link} \\
\midrule
MiniCPM-o-2.6  &  8B & \url{https://huggingface.co/openbmb/MiniCPM-o-2_6} \\
Ultravox-v0.5-llama-3.1-8b  &  8B & \url{https://huggingface.co/fixie-ai/ultravox-v0_5-llama-3_1-8b} \\
Ultravox-v0.6-Llama-3.1-8b &  8B &  \url{https://huggingface.co/fixie-ai/ultravox-v0_6-llama-3_1-8b}\\
Phi-4-Multimodal-Instruct  & 5B & \url{https://huggingface.co/microsoft/Phi-4-multimodal-instruct} \\
R1-AQA & 7B & \url{https://huggingface.co/mispeech/r1-aqa} \\

SeaLLMs-Audio-7B & 7B & \url{https://huggingface.co/SeaLLMs/SeaLLMs-Audio-7B} \\

Qwen2-Audio-7B-Instruct  & 7B & \url{https://huggingface.co/Qwen/Qwen2-Audio-7B-Instruct} \\
Qwen2.5-Omni-3B  & 3B & \url{https://huggingface.co/Qwen/Qwen2.5-Omni-3B} \\

Qwen2.5-Omni-7B  & 7B & \url{https://huggingface.co/Qwen/Qwen2.5-Omni-7B} \\
Qwen3-Omni-30B-Instruct  &  30B & \url{https://huggingface.co/Qwen/Qwen3-Omni-30B-A3B-Instruct}\\
GPT-4o-Audio-Preview  &  -- & \url{https://platform.openai.com/docs/models/gpt-4o} \\
Gemini-2.5-Flash  & --  & \url{https://deepmind.google/technologies/gemini/} \\
Gemini-3-Flash  & --  & \url{https://deepmind.google/technologies/gemini/} \\
Gemini-3-Pro  &  -- & \url{https://deepmind.google/technologies/gemini/} \\
\bottomrule
\end{tabular}
\label{tab:model_names}
\end{table*}

\label{SupEnd}
\end{document}